\documentclass[10pt,twocolumn,letterpaper]{article}

\usepackage[pagenumbers]{wacv}

\usepackage{graphicx}
\usepackage{amsmath}
\usepackage{amssymb}
\usepackage{booktabs}

\usepackage[pagebackref,breaklinks,colorlinks]{hyperref}

\usepackage[capitalize]{cleveref}
\crefname{section}{Sec.}{Secs.}
\Crefname{section}{Section}{Sections}
\Crefname{table}{Table}{Tables}
\crefname{table}{Tab.}{Tabs.}

\usepackage{listings}
\usepackage{color} %
\usepackage{colortbl,array,xcolor}
\usepackage{float}
\usepackage{enumitem}

\usepackage{multirow}

\newfloat{lstfloat}{htbp}{lop}
\floatname{lstfloat}{Listing}

\definecolor{codegreen}{rgb}{0,0.6,0}
\definecolor{codegray}{rgb}{0.5,0.5,0.5}
\definecolor{codepurple}{rgb}{0.58,0,0.82}
\definecolor{backcolour}{rgb}{0.95,0.95,0.92}

\def\wacvPaperID{924} %
\def\confName{WACV}
\def\confYear{2025}

\begin{document}

\title{Advancing Large Multi-modal Models with Explicit Chain-of-Reasoning and Visual Question Generation}

\author{Kohei Uehara$^{1,\,2}$
\and
Nabarun Goswami$^{1,\,2}$
\and
Hanqin Wang$^{1}$
\and
Toshiaki Baba$^{1,\,2}$
\and
Kohtaro Tanaka$^{1,\,2}$
\and
Tomohiro Hashimoto$^{1,\,2}$
\and
Kai Wang$^{1,\,2}$
\and
Rei Ito$^{1}$
\and
Takagi Naoya$^{1,\,2}$
\and
Ryo Umagami$^{1}$
\and
Yingyi Wen$^{1}$
\and
Tanachai Anakewat$^{1,\,2}$
\and
Tatsuya Harada$^{1,\,2}$\\
\and
$^{1}$The University of Tokyo, $^{2}$RIKEN
\and
{\tt\small \{uehara, nabarungoswami, wang, baba, k-tanaka, hashimoto,}\\{\tt\small  wang-kai, ito, takagi, umagami, wenyy, anakewat, harada\}@mi.t.u-tokyo.ac.jp}
}
\maketitle

\begin{abstract}
The increasing demand for intelligent systems capable of interpreting and reasoning about visual content requires the development of large Vision-and-Language Models (VLMs) that are not only accurate but also have explicit reasoning capabilities.
This paper presents a novel approach to develop a VLM with the ability to conduct explicit reasoning based on visual content and textual instructions.
We introduce a system that can ask a question to acquire necessary knowledge, thereby enhancing the robustness and explicability of the reasoning process.
To this end, we developed a novel dataset generated by a Large Language Model (LLM), designed to promote chain-of-thought reasoning combined with a question-asking mechanism.
The dataset covers a range of tasks, from common ones like caption generation to specialized VQA tasks that require expert knowledge.
Furthermore, using the dataset we created, we fine-tuned an existing VLM.
This training enabled the models to generate questions and perform iterative reasoning during inference.
The results demonstrated a stride toward a more robust, accurate, and interpretable VLM, capable of reasoning explicitly and seeking information proactively when confronted with ambiguous visual input.
\end{abstract}
\section{Introduction}\label{sec:intro}

\begin{figure}
    \centering
    \includegraphics[width=\linewidth]{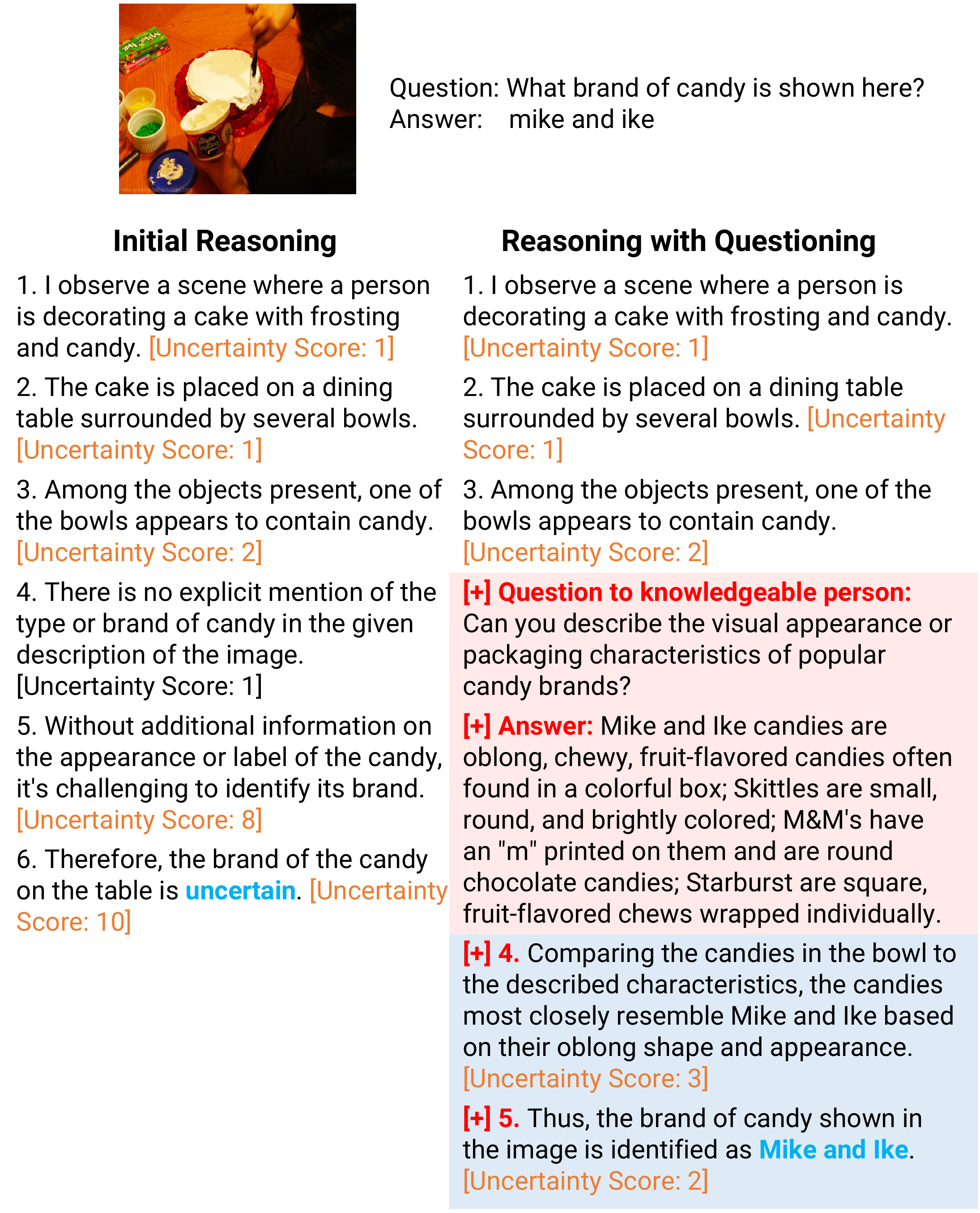}
    \caption{An example of the explicit reasoning steps we aim to achieve, also representing a sample from our constructed dataset.
    It demonstrates the thought process in response to a given question.
    Notably, we incorporate a question generation step into the reasoning process (as seen on the right side of the figure), allowing the model to interactively acquire knowledge and refine its reasoning steps.}
    \label{fig:intro}
\end{figure}

In recent years, LLMs have gained significant popularity in the field of artificial intelligence~\cite{gpt3,OpenAI2023GPT4TR,Touvron2023Llama2O}.
Building upon this, research has steped into large Vision-and-Language Models (VLMs)~\cite{Li2023BLIP2BL,dai2023instructblip,liu2023llava,liu2023improvedllava}.
Our study specifically focuses on tasks involving both vision and language modalities.
Generally, VLMs learn the alignment between images and texts using large datasets sourced by web crawling.
They are then fine-tuned with smaller, high-quality datasets for instruction tuning to enhance their text generation capabilities for various tasks.
However, these models often struggle with hallucination, where they produce outputs not aligned with the given input.
For instance, in VLMs, the models may mention objects that do not exist in the given images~\cite{liu2023aligning,liu2023hallusionbench}.

A critical limitation of these models is their inability to explain the reasoning behind their outputs, especially when hallucination occurs.
This is a significant drawback, as it is difficult to identify and correct the model's mistakes.
We argue that these issues stem from the models being trained to directly output answers to instructions without including a reasoning process.
By explicitly learning and outputting the reasoning process, models can provide more concrete and explanatory responses.
Another limitation is that the models cannot correct their mistakes by themselves, as they lack the ability to reason about their own outputs.
On the other hand, humans, when uncertain, often ask questions to acquire new knowledge and refine their answers.
By explicitly outputting the reasoning process, models can be trained to generate questions during the reasoning steps, pause to acquire necessary knowledge from knowledgeable someone, and then continue the reasoning process, thereby arriving at more accurate and reliable conclusions.

Therefore, our research aims to address these challenges by incorporating an explicit reasoning process and the ability to generate questions during reasoning.
This approach is akin to the Chain-of-Thought reasoning seen in LLMs~\cite{wei2022chain,chuCoTReasoningSurvey2023}, where models are prompted to explicitly reason out problems.
However, due to VLMs' relatively weaker text generation capabilities compared to LLMs, achieving Chain-of-Thought solely through prompting is challenging.
To overcome this, we have created a novel dataset that includes the explicit reasoning process.
Following mainstream methods in LLM-based dataset creation~\cite{liu2023llava}, we generate this data using LLMs, combining image annotations and a few manually created examples.
This data includes scenarios where the model needs to generate questions when uncertain, training it to ask questions during ambiguous reasoning situations.
We refer to this reasoning process as ``Chain-of-Reasoning (CoR).''

In order to present the effectiveness of our approach, we fine-tune an existing VLM model on the dataset we created.
Being trained on our dataset, the model acquires the ability to generate explicit reasoning steps and ask questions when uncertain, thereby improving the reliability of its inferences.
Our contributions are summarized as follows:
\begin{itemize}
  \item We present an approach by incorporating an explicit reasoning process and question-generation capability into VLMs, promoting more reliable inferences.
  \item We crafted a new dataset and utilized it for model training, setting a precedent for future VLM advancements.
  \item With our novel dataset and model, we achieved giving the model the ability to generate explicit reasoning steps and question-asking capability.
\end{itemize}

\section{Related Work}\label{sec:relatedwork}

\subsection{Large-scale Vision-and-Language Models}\label{sec:relatedwork:lmm}

The development of LLMs like ChatGPT~\cite{chatgpt}, GPT-4~\cite{OpenAI2023GPT4TR}, and LLaMA~\cite{Touvron2023Llama2O} has paved the way for research into large-scale multi-modal models that extend beyond text-based modalities.
This section highlights studies particularly relevant to our research, focusing on models integrating vision and language modalities.

BLIP-2~\cite{Li2023BLIP2BL} and InstructBLIP~\cite{dai2023instructblip} utilize pre-trained image encoders and text decoders, employing a Transformer model known as Q-Former as an adapter.
The training of these models is typically divided into two stages.
In the first stage, Q-Former learns the alignment between image and text using large-scale image-text paired datasets.
The second stage involves training the Q-Former alongside the text decoder to enable the decoder to produce outputs according to given instructions.
Generally, while the first training stage prioritizes the scale of the dataset, the second stage emphasizes the quality of smaller, high-quality instruction and answer datasets. 
This two-stage learning approach is widely adopted in the training of VLMs, as seen in models like BLIVA~\cite{hu2023bliva}, MiniGPT (v1\&2)~\cite{zhu2023minigpt,chen2023minigptv2}, and LLaVA (v1\&1.5)~\cite{liu2023llava,liu2023improvedllava}.

Additionally, there's growing interest in VLMs that not only produce outputs aligned with overall image instructions but also focus on specific regions within images.
This research direction is evident in models such as KOSMOS-2~\cite{kosmos2}, Shikra~\cite{chen2023shikra}, MiniGPT v2~\cite{chen2023minigptv2}, and LLaVA v1.5~\cite{liu2023improvedllava}.
These models transform bounding box coordinates, representing input regions, into special tokens for textual representation.
Another noteworthy model is GPT4RoI~\cite{zhang2023gpt4roi}, which applies an RoIAlign based on the target region to the features outputted by the image encoder, thus obtaining features focused on the specified region.

\begin{figure*}[t]
    \centering
    \includegraphics[width=0.9\linewidth]{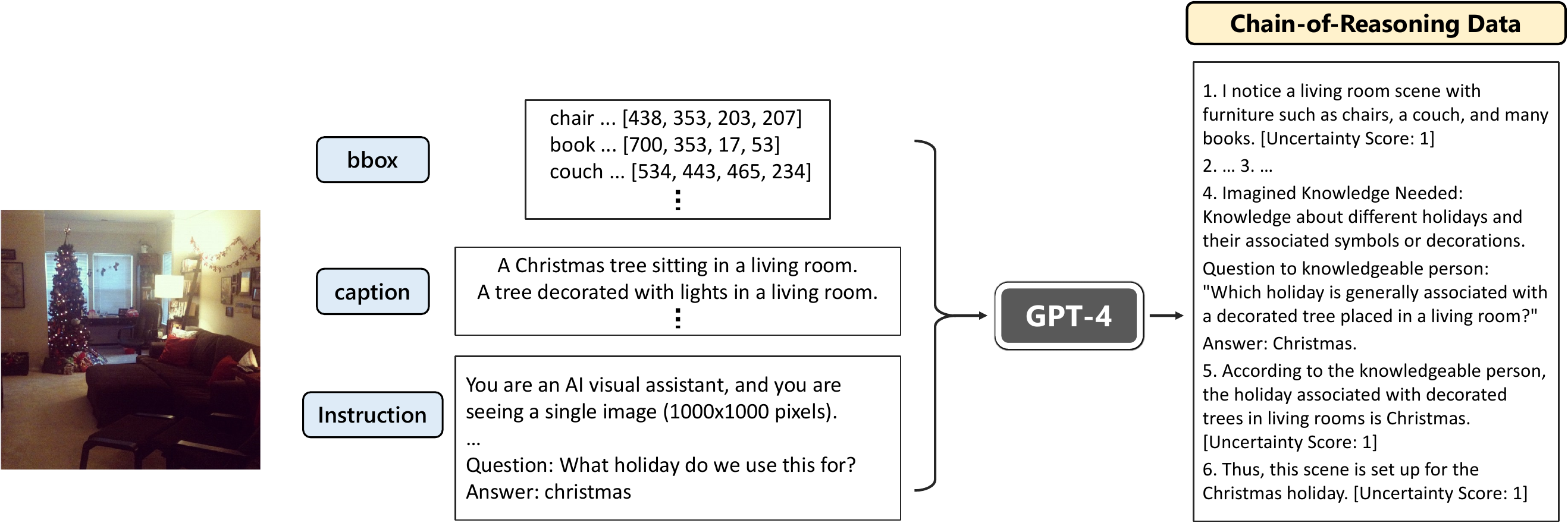}
    \caption{
        An overview of the dataset construction process.
        The LLM takes bounding boxes information, image captions, and instructions as input and generates reasoning steps and questions as output.
    }
    \label{fig:dataset}
\end{figure*}

\begin{figure}
    \centering
    \includegraphics[width=0.9\linewidth]{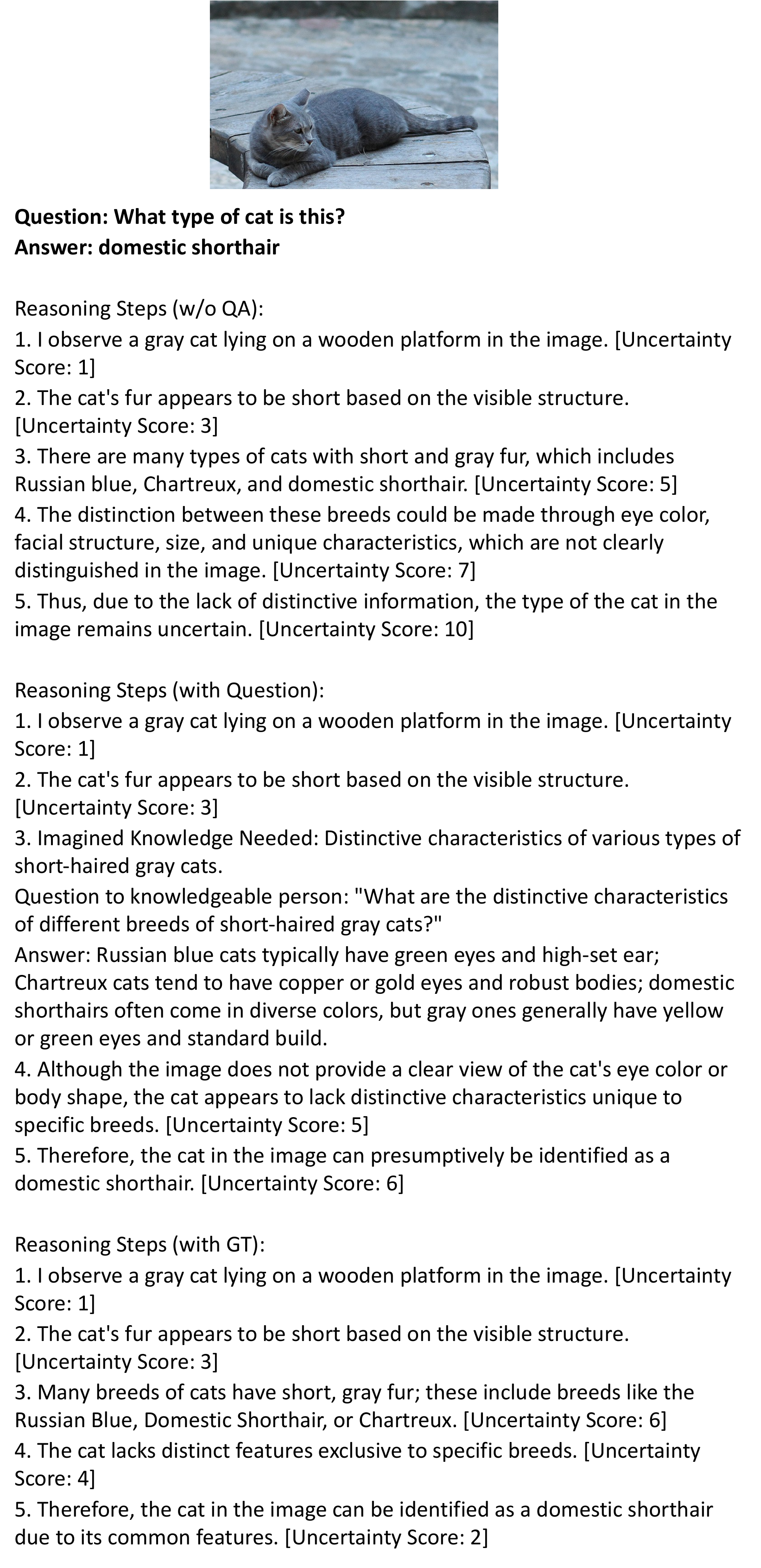}
    \caption{
        An example from our dataset, created from the OK-VQA dataset.
        To the given question, the dataset contains a series of reasoning steps in three settings: without QA, with QA, and with GT.
    }
    \label{fig:data_example}
\end{figure}

\subsection{Explicit Reasoning in V\&L Tasks}\label{sec:relatedwork:er}

In the field of natural language processing tasks, LLMs have shown the capability to perform explicit reasoning in a format known as Chain-of-Thought~\cite{chuCoTReasoningSurvey2023}, thanks to their advanced language generation abilities.
However, in vision-and-language (V\&L) tasks, the adoption of explicit reasoning using large-scale models is still in its developmental phase compared to language-only tasks.
The closest research in this area involves using LLMs to predict necessary reasoning steps in a programmatic format and subsequently calling APIs specialized for V\&L tasks, such as in Visual Programming~\cite{Gupta_2023_CVPR} and ViperGPT~\cite{surismenon2023vipergpt}.
A significant limitation of these approaches is that the LLM predicting the reasoning steps does not have direct access to the images, leading to a lack of assurance that the reasoning is genuinely based on the image content.
Our model addresses this gap by allowing the V\&L model to directly predict reasoning steps, ensuring that the reasoning is firmly grounded in the image content.

\subsection{Visual Question Generation}\label{sec:relatedwork:vqg}

A related field to the concept of acquiring information through question generation is Visual Question Generation (VQG).
In the early stages of VQG, the primary focus was on generating questions related to images, without any specific goal of acquiring knowledge~\cite{mostafazadehVQG,iqan,ivqa,vqg_unknown}.
K-VQG~\cite{Uehara_2023_WACV,uehara2024learning} does involve generating questions for knowledge acquisition; however, there has been no emphasis on utilizing the acquired knowledge in further reasoning steps.

Learning by Asking (LBA) is another relevant area.
LBA is an approach where question generation is used to acquire data for learning purposes.
For instance, LBA has been employed in tasks like VQA~\cite{Misra_2018_CVPR,Uehara_2022_CVPR,uehara2024learning}, caption generation~\cite{shen2018learning}, and scene graph completion~\cite{pmlr-v87-yang18a}.
However, our research significantly diverges from LBA, as we focus on generating questions as part of explicit reasoning during the inference process.
The acquired answers are then immediately used to update and refine the reasoning quality, differentiating our approach from traditional LBA work.
\section{Method}\label{sec:method}
First, we introduce how we constructed our CoR dataset.
Then, we explain the model details and training process.

\subsection{Dataset Construction}\label{sec:method:dataset}
\begin{lstfloat}[t]
\begin{lstlisting}[label={lst:prompt},language={},caption={Full example of our prompt given to GPT-4.}]
You are an AI visual assistant, and you are seeing a single image (1000x1000 pixels). What you see are provided with captions, describing the same image you are looking at, and the coordinates of the bounding box of objects in the image. Provide the detailed reasoning steps for the given question as you are seeing the image. Please consider you cannot access the answer, but you can use a knowledgeable person to ask a question.
You need to generate reasoning steps in three settings:
1. You are not allowed to ask questions to a knowledgeable person. In this case, you may not reach the correct answer in the end.
2. You can ask questions to a knowledgeable person to gain additional knowledge. You must keep the same reasoning steps as the case 1,  until you ask questions.
3. You are not allowed to ask questions, but you can see the answer to the question. So, you must construct your reasoning step to definitely reach the answer.
Note that the captions and bounding boxes are provided as a reference, so you need to behave like you are seeing the image.
IMPORTANT: The presence of these captions or descriptions,  bboxes, and the answer must not be included in reasoning steps. Behave like you are actually seeing the image.
Thus, you must not use the word "caption", "description", or "bounding box" in the reasoning steps.
Especially, do not mention the coordinates of each object in the image. If you want to mention the position of the object, please refer to it by relative position, such as "at the left of the image" or "in front of xx".
Please provide reasoning steps only, do not include other sentences or contents.
The reasoning step should include only the information necessary to infer the answer to the question, and should not refer to anything that is not directly related to the question or answer.
Each reasoning step must contain only one sentence, and the total size of the reasoning steps must not be over seven.
Please add an uncertainty score [1-10, higher is more uncertain] to each reasoning step.
Avoid leaps of logic between each step, but if there are any, please increase the uncertainty score in that step.
IMPORTANT: In case 2, if there is any chance to decrease the uncertain score by asking questions to other knowledgeable people, you must ask questions to obtain additional information. Before asking the question, you need to imagine what kind of knowledge is needed to decrease the uncertainty, and the question must be one that you can get the knowledge. The question must not be a paraphrase of the given question. Include a specific statement of your imagined knowledge, question, and the answer to your question. Please continue the reasoning as if you could get the answer to the questions. Note that the question and the answer should not be counted as the reasoning steps.
\end{lstlisting}
\end{lstfloat}

\begin{figure*}
    \centering
    \includegraphics[width=0.7\linewidth]{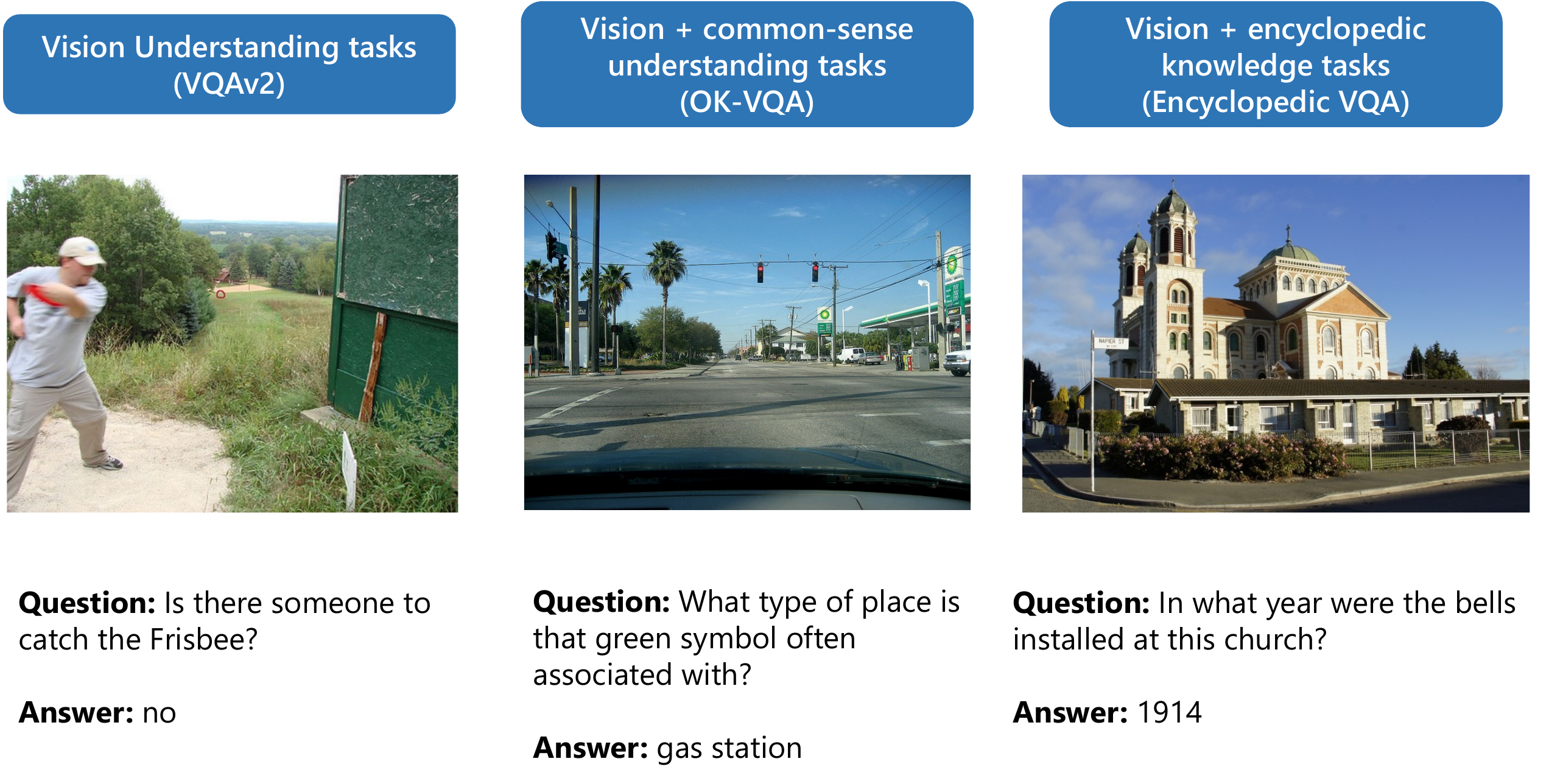}
    \caption{
        Examples of dataset in each category: visual understanding tasks, vision + common-sense understanding tasks, and vision + encyclopedic knowledge tasks.
    }
    \label{fig:category}
\end{figure*}

In pursuit of realizing the CoR, we developed a novel dataset.
This dataset was created using a combination of the specially designed prompt, manually curated examples and the text generation capabilities of LLM.
We show an overview of the dataset construction process in Figure~\ref{fig:dataset} and an example of the dataset in Figure~\ref{fig:data_example}.

The data structure of our dataset consists of the image, an instruction text of the task, a question (optional, if the task is in QA format), the answer, and the reasoning process required to derive the answer.
We used different approaches to build the dataset depending on the types of annotations provided with the base dataset. 

(1) If the dataset had rich annotations, including captions and object detection annotations, such as labels and bounding boxes, we followed the LLaVA approach~\cite{liu2023llava}.
In this case, we used an LLM (GPT-4~\cite{achiam2023gpt}) that does not process images to build the dataset.
Coupling these inputs with thoughtfully designed prompts (Listing~\ref{lst:prompt}) enables the LLM to generate the appropriate reasoning steps.
It's crucial that the prompts ensure that the LLM simulates the act of observing the image directly, without mentioning any references to captions, descriptions, or bounding boxes.
This approach ensures the consistency of the reasoning process when training VLMs, which do view the image directly.
To achieve this, in the prompt given to the LLM, we clearly state like \textit{you must not use the word ``caption'', ``description'', or ``bounding box'' in the reasoning steps.
Especially, do not mention the coordinates of each object in the image.
If you want to mention the position of the object, please refer to it by relative position, such as ``at the left of the image'' or ``in front of xx''.}
Furthermore, in the prompt, we instruct the LLM to suppose that the image is of size 1000$\times$1000 pixels.
This ensures accurate position estimations based on given textual coordinates of the bounding boxes.
The prompts also emphasize succinct reasoning steps without redundancy, and the need to add \textit{uncertainty scores} at the end of each reasoning step.

(2) If the dataset did not have such rich annotations, we used a model that can input images (GPT-4V) to build the dataset.
We used the similar prompts as previous case, excluding the parts that required GPT to imagine image input from text information.

We devised three variants of reasoning step data: without QA, with QA, and with GT.
In ``without QA'' setting, the reasoning process to the answer is conducted without generating any questions midway.
This setting aims to mimic the process a VLM undergoes when attempting to answer a question directly, without generating any questions.
In this scenario, the LLM is instructed to generate reasoning steps without seeing the correct answer to the given task.
We instruct the model to generate ``Uncertainty score'' along with each reasoning step, indicating the model's confidence in its reasoning.
This plays a crucial role in ``with QA'' setting described below.

In contrast, the ``with QA'' setting incorporates the generation of questions as part of the reasoning steps.
This setup is crucial for training VLMs to perform explicit reasoning while simultaneously generating questions.
As the model generates the reasoning steps, it also generates an uncertainty score for each step.
If the score significantly rises at any point in the ``without QA'' setting, it triggers the generation of a question just before the step.
This mechanism helps the model to identify moments of uncertainty and seek external information, thereby simulating a more interactive and dynamic reasoning process.

Lastly, the ``with GT'' setting involves the LLM generating reasoning steps while having access to the correct answers to the tasks (i.e., the ground-truth answers for the given questions or captions for the given images).
This approach differs from the previous two as the reasoning steps invariably lead to the correct answer, serving as a ground truth for correct reasoning without question generation.
This data is primarily used to train VLMs that do not involve question generation in their reasoning process.

\begin{table*}[]
    \centering
    \begin{tabular}{@{}lccccc@{}}
    \toprule
     &  &  & \multicolumn{3}{c}{Average number of steps} \\
     & Num. of samples & Num. of images & without QA & with QA & with GT \\ \midrule
    COCO Caption & 5,857 & 5,782 & 6.27 & 8.75 & 6.20 \\
    VQA v2 & 5,755 & 5,633 & 4.68 & 7.36 & 4.54 \\
    OK-VQA & 5,793 & 5,792 & 4.77 & 7.35 & 4.55 \\
    A-OKVQA & 5,736 & 5,718 & 4.87 & 7.43 & 4.63 \\
    Visual Genome & 5,883 & 5,609 & 4.34 & 7.31 & 4.12 \\
    Encyclopedic VQA & 6,521 & 6,186 & 7.00 & 9.45 & 7.00 \\
    OVEN & 5,685 & 5,685 & 6.99 & 9.51 & 6.99 \\ \midrule
    Total & 41,230 & 39,272 & 5.58 & 8.19 & 5.46 \\ \bottomrule
    \end{tabular}
    \caption{
        Dataset statistics of our dataset.
        In each row, we show the statistics of the base dataset to craft our dataset.
        ``Num. of images'' indicates the number of unique images in the dataset.
        ``without QA'', ``with QA'', and ``with GT'' indicate the average number of steps when the dataset is generated without questions and answers, with questions and answers, and without seeing ground-truth answers, respectively.
    }
    \label{tab:dataset-stats}
\end{table*}

Each question step is designed to consist of three key elements: \textbf{Imagined Knowledge Needed}, the \textbf{Question}, and the \textbf{Answer}.
Firstly, ``Imagined Knowledge Needed'' component represents the knowledge the model predicts it needs to acquire during the reasoning process.
Let us consider the example shown in Figure~\ref{fig:dataset}.
Here, the model need to answer the question ``What holiday do we use this for?'', according to the given image of Christmas decorations.
The model might need more specific information about different holidays and their associated symbols or decorations.

Secondly, we include the ``Question'' that would facilitate the acquisition of this knowledge.
For instance, the model might generate a question to acquire imagined knowledge needed, such as ``Which holiday is generally associated with a decorated tree placed in a living room?''
This step is crucial as it reflects the model's ability to formulate relevant questions based on the required knowledge.

The third component is the ``Answer'', which is essentially the answer to the generated question.
For example, the answer to the question above would be ``Christmas'', and then the model is able to successfully acquire the knowledge it needs to answer the original question.

\subsection{Dataset Statistics}\label{sec:method:dataset:stats}

\begin{figure*}
    \centering
    \includegraphics[width=0.7\linewidth]{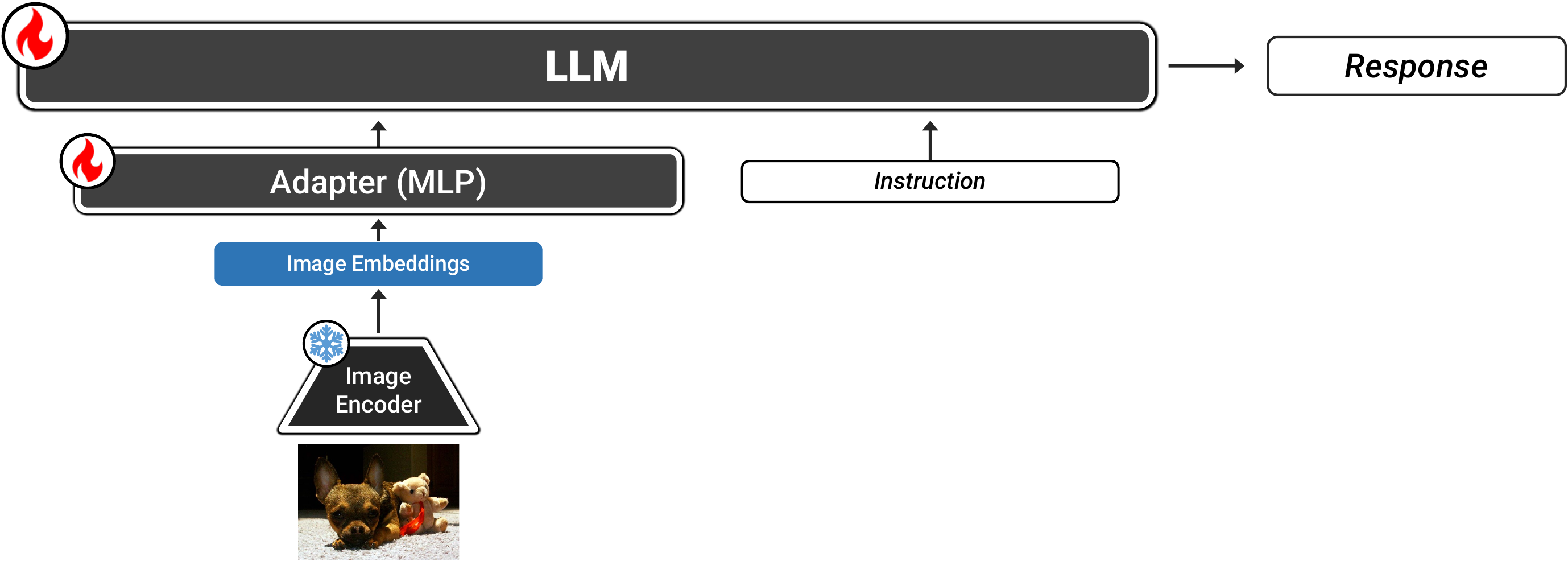}
    \caption{
        An overview of the model.
        Image encoders extract embeddings from input images, which are fed into the Adapter MLPs.
        The extracted image feature and instruction texts are fed into the LLM, culminating in the generation of a text response.
        This architecture enables the model to consider both visual information and textual instructions in its reasoning process.
    }
    \label{fig:model}
\end{figure*}

We provide a comprehensive overview of our dataset statistics in Table~\ref{tab:dataset-stats}.
In our research, we based our dataset generation on seven datasets: COCO Caption~\cite{Lin2014MicrosoftCC}, Visual Genome Caption~\cite{krishna2017visual}, VQA v2~\cite{Goyal2016MakingTV}, OK-VQA~\cite{okvqa}, A-OKVQA~\cite{Schwenk2022AOKVQAAB}, OVEN~\cite{hu2023open}, and Encyclopedic-VQA~\cite{Mensink_2023_ICCV}.
We can categorize these datasets into three groups:\\
\textbf{(1) Visual understanding tasks.}\noindent \\
This group includes COCO Caption, Visual Genome Caption, and VQA v2, which are captioning tasks and VQA tasks that require the model to understand or describe the content of the image.
Usually, these tasks do not require the model to have expert knowledge that cannot be directly observed from the image.\\
\textbf{(2) Vision + common-sense understanding tasks.}\\
This group includes OK-VQA and A-OKVQA, which require the model to have common-sense knowledge to answer the questions.
Here, common-sense knowledge refers to general knowledge about topics such as geography, brands, and vehicles.~\cite{okvqa}\\
\textbf{(3) Vision + encyclopedic knowledge tasks.}\\
This group includes Encyclopedic-VQA and OVEN, which require the model to have encyclopedic knowledge to answer the questions.
Encyclopedic knowledge refers to more in-depth and specific knowledge compared to common-sense knowledge.
This includes knowledge such as the names of animal and plant species, or details about when and by whom a building in an image was constructed~\cite{Mensink_2023_ICCV}.

We assume that the difficulty of the reasoning process increases as we move from visual understanding tasks to vision + encyclopedic knowledge tasks, as the latter requires detailed and expert knowledge to answer the questions.
For each of these datasets, we generated approximately 6,000 data samples, randomly selecting from their training sets.
We show examples of each category in Figure~\ref{fig:category}.

From Table~\ref{tab:dataset-stats}, we can see that when generating reasoning data without questions and answers (without QA) and when not seeing ground-truth answers (with GT), the average number of reasoning steps is shorter compared to the scenario where both questions and answers are included (with QA).
This is attributed to the fact that the QA generation process inherently requires more reasoning steps.

The ``with GT'' setting shows slightly fewer reasoning steps than the without QA setting.
This can be attributed to providing the correct answers to GPT-4, which likely allows for concise reasoning steps leading to the correct answer.

The COCO Caption, Encyclopedic-VQA, and OVEN dataset tend to have longer reasoning steps compared to the others.
For COCO Caption, this is likely due to the nature of the task, which involves generating captions based on the entire content of the image, requiring a more comprehensive reasoning process.
The Encyclopedic-VQA and OVEN datasets often contains questions that require very detailed observations or expert knowledge to answer, leading to longer reasoning steps.
In contrast, the other tasks focus on specific questions or target regions within the images, thereby requiring less information to be incorporated into the reasoning steps.

\subsection{Model Architecture}\label{sec:method:model}

Our model fundamentally builds upon the LLaVA architecture~\cite{liu2023llava}, consisting of an image encoder, a text decoder, and an adapter component based on two-layer MLP blocks.
We show an overview of our model in Figure~\ref{fig:model}.
The input image is first resized to 336$\times$336 pixels and passed through a pre-trained image encoder (e.g., CLIP-ViT~\cite{pmlr-v139-radford21a}).
The image encoder extracts the image features, which are then fed into the Adapter MLPs to transform the image features into a format that can be processed by the LLM.
The instruction text is also fed into the LLM, and the model generates a text response based on the image features and instruction text.

\begin{table*}[]
    \centering
\begin{tabular}{@{}lcccccccc@{}}
\toprule
 & \begin{tabular}[c]{@{}c@{}}\footnotesize{COCO}\\\footnotesize{Caption}\end{tabular} & \begin{tabular}[c]{@{}c@{}}\footnotesize{Visual}\\ \footnotesize{Genome}\end{tabular} & \footnotesize{OK-VQA} & \footnotesize{A-OKVQA} & \footnotesize{VQA v2} & \begin{tabular}[c]{@{}c@{}}\footnotesize{Encyclopedic}\\ \footnotesize{VQA}\end{tabular} & \footnotesize{OVEN} & \footnotesize{average} \\ \midrule
LLaVA (original) & \textbf{2.258} & 1.424 & 2.414 & 2.332 & 2.604 & 1.606 & 1.521 & 2.023 \\
CoR \footnotesize{w/o question} & 1.621 & 1.653 & 2.224 & 2.054 & 2.524 & 1.520 & 1.582 & 1.883 \\
CoR \footnotesize{w/ uncertainty} & 1.748 & 1.748 & 2.536 & 2.408 & 2.715 & 1.902 & 1.761 & 2.117 \\
CoR \footnotesize{w/ uncertainty, w/o knowledge} & 1.775 & 1.739 & 2.526 & 2.376 & 2.717 & 1.909 & \textbf{1.839} & 2.126 \\
\textbf{Ours CoR} & 1.769 & \textbf{1.782} & \textbf{2.631} & \textbf{2.459} & \textbf{2.737} & \textbf{1.925} & 1.836 & \textbf{2.163} \\ \bottomrule
\end{tabular}
\caption{
    Evaluation results of our model on various datasets.
    The evaluation is conducted using GPT-4 to score the reasoning steps generated by the model.
    The scores range from 1 to 4, with 4 indicating a correct answer.
    The average score across all datasets is shown in the last column.
}
\label{tab:results}
\end{table*}

\subsection{Training}\label{sec:training}

We utilizes our dataset to make the model learn to generate reasoning steps and questions.
In the training process, we utilized a pre-trained VLM, which is already instruction-tuned, and fine-tuned it on our dataset.
In the fine-tuning process, we froze the parameters of the image encoder, and updated the parameters of the text decoder and the adapter MLP.
Along with the image and original question, we provided a specially designed prompts, such as ``\textit{Analyze the image and outline your reasoning process step by step before providing your final answer},'' to let the model generate reasoning steps and questions.

When the model is applied to test data, it performs a two-stage reasoning process to reach the final answer.
The first stage encompasses the reasoning that leads up to the generation of a question.
The second stage occurs after an external response to the question is obtained; the model is given the original inputs, the intermediate reasoning steps, and the external response.
Then, the model generates the remaining reasoning steps based on this new information.

In this study, we leverage GPT-4o as the external answerer.
It is noteworthy that the questions generated by the model typically demand an understanding of the image content as well as common-sense and encyclopedic knowledge.
Given its extensive training data and large model parameters, GPT-4 is expected to adeptly handle the knowledge required to answer the questions.

\section{Experiments}\label{sec:experiments}

\subsection{Implementation Details}\label{sec:experiments:implementation}

Our experimental setup incorporates the following models: for image processing, we use CLIP-ViT-Large\footnote{https://huggingface.co/openai/clip-vit-large-patch14-336}, and for text generation, we employ vicuna-7b-1.5\footnote{https://huggingface.co/lmsys/vicuna-7b-v1.5}.
The entire model is trained from a publicly available pre-trained checkpoint~\footnote{https://huggingface.co/llava-hf/llava-1.5-7b-hf}.

In fine-tuning stage, we utilized 8 $\times$ NVIDIA H100 (80GB) GPU for about 1 hour.
We trained the model for one epoch with a total batch size of 128.
The learning rate was set to a maximum of 2e-5, linearly increasing until 3\% of the training steps, followed by a cosine decay schedule.
Training was conducted using bfloat16 precision.

\begin{figure*}[t]
    \centering
    \includegraphics[width=0.9\linewidth]{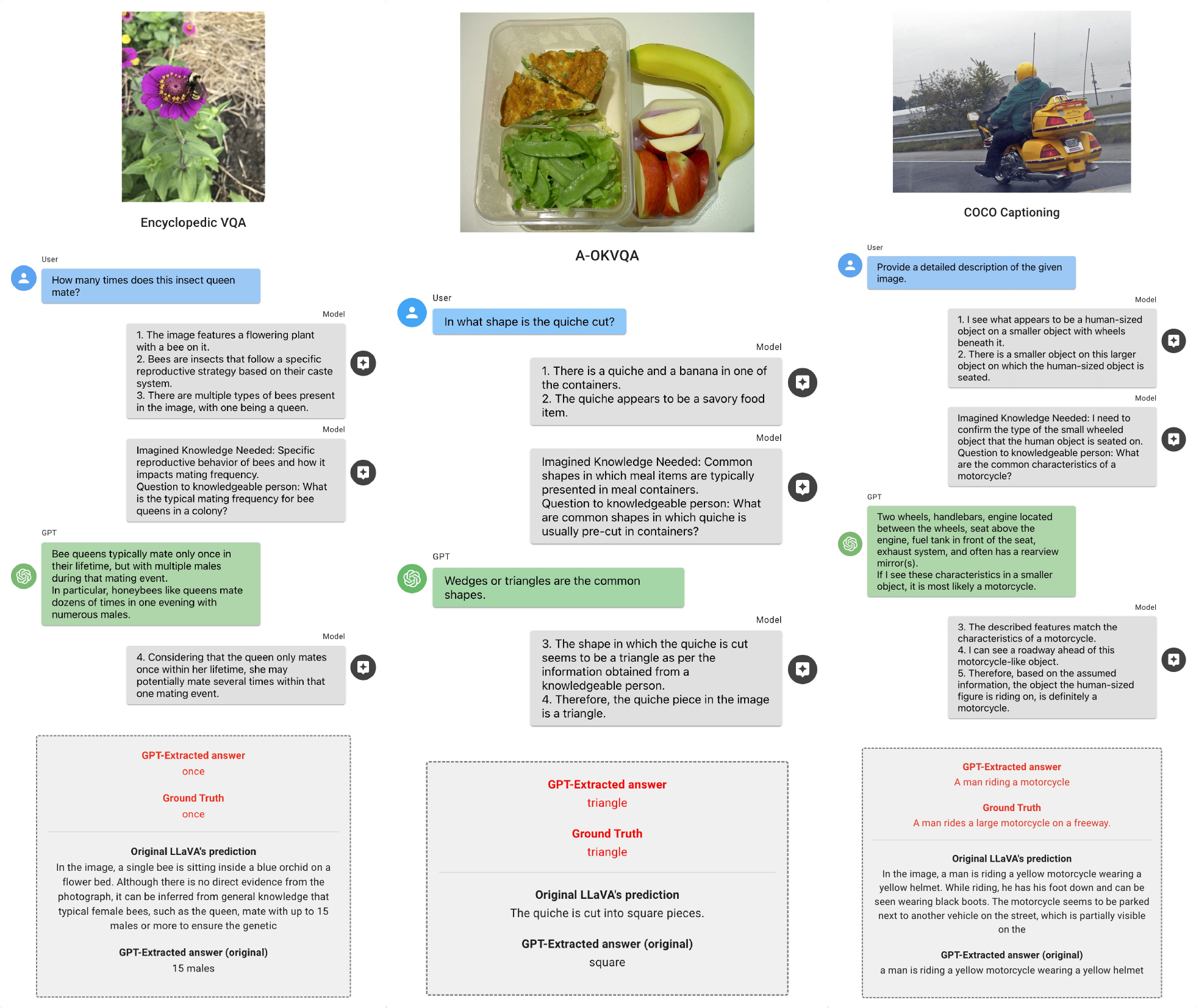}
    \caption{
        An example of the predicted reasoning steps for Encyclopedic-VQA, A-OKVQA, COCO Captioning, which are a representative sample from our constructed dataset's category.
    }
    \label{fig:result}
\end{figure*}

\subsection{Evaluation Settings}\label{sec:experiments:evaluation}
In our experiments, we investigated how the presence or absence of CoR affects the behavior of the model when solving tasks.
We also evaluated several ablations related to the CoR method.
These included listing the necessary knowledge before question generation and generating an uncertainty score along with the reasoning steps.
We used validation set of seven datasets for this evaluation: MS COCO, Visual Genome, VQAv2, OK-VQA, A-OKVQA, OVEN, and Encyclopedic VQA.
Using entire validation set is computationally expensive, so we randomly sampled 1000 instances from each dataset for evaluation.
Our experiments compared the model trained with CoR to a baseline model trained without question generation, specifically using the ``with GT'' setting from the dataset (referred to as ``w/o Question'').

Currently, there are no established metrics for evaluating results that include long reasoning steps.
Therefore, in this study, we used GPT-4 for scoring the answers.
We instructed GPT-4 to extract the final answer from the reasoning steps and then evaluate the answer with a score from 1 to 4 (1: incorrect, 2: partially correct, 3: mostly correct, 4: correct).

\subsection{Results and Discussions}\label{sec:experiments:results}
The results of our experiments are shown in Table~\ref{tab:results}.
Also, we show an example of the predicted reasoning steps in Figure~\ref{fig:result}.

The evaluation results show that ``Ours CoR'' performs the best on average across various datasets.
In contrast, the CoR model without questions (CoR w/o question) has the lowest performance.
This might be because the current model cannot consistently provide coherent outputs over long reasoning processes.
The challenge of developing VLMs that can produce coherent and consistent long reasoning steps in line with the given tasks remains an important area for future research.

Comparing ``Ours CoR'' with the original LLaVA, the latter performs relatively well on VQA v2 and COCO Caption, which are part of LLaVA’s training data.However, for datasets not included in LLaVA’s training data, like OVEN and Encyclopedic VQA, which require specialized knowledge, models that ask questions (such as ``Ours CoR'') show significantly better performance.
This indicates that asking questions helps the model acquire necessary specialized knowledge that it does not inherently possess. 

Furthermore, outputting uncertainty did not significantly improve performance (CoR w/ uncertainty vs. Ours CoR).
This is likely because generating uncertainty outputs makes the model's responses longer and less consistent, leading to performance degradation.
However, if the model can handle uncertainty effectively, it could ask questions during the most uncertain reasoning steps, potentially improving performance.

Comparing models with and without prior knowledge output (CoR w/o knowledge vs. Ours CoR), it's evident that pre-outputting necessary knowledge improves performance.
This prevents the model from asking irrelevant questions and ensures that it focuses on helpful inquiries, leading to better results.
\section{Conclusion}\label{sec:conclusion}

In this paper, we have presented a novel approach of improving the capabilities of VLMs.
This was achieved by incorporating a structured CoR and the ability to generate questions during the reasoning process.
The model was trained on a dataset specifically designed to include an explicit reasoning and question-asking process.

We proposed an architecture for a VLM that leverages pre-trained components such as an image encoder and text decoder, all of which are fine-tuned using our novel dataset.
Our experimental results suggest that the model's ability to generate questions contributes to its performance, compared to a baseline model that does not generate questions.
This underscores the potential utility of integrating explicit reasoning processes into VLMs.

\section*{Acknowledgements}
This work was partially supported by JST Moonshot R\&D Grant Number JPMJPS2011, CREST Grant Number JPMJCR2015 and Basic Research Grant (Super AI) of Institute for AI and Beyond of the University of Tokyo.

\setcounter{section}{0}
\renewcommand{\thesection}{\Alph{section}}
\clearpage
\section*{Appendix}
\subsection*{A. Prompt Details}
\begin{lstfloat}[h]
\begin{lstlisting}[label={lst:prompt_answer},language={},caption={Full example of our prompt given to GPT-4 to answer the question generated by our model.}]
You are a helpful assistant to a user who is trying to answer the question about an image.
The user has provided a question and a reasoning process that led to their answer.
The user asks you a question to collect knowledge that will help them answer the question.
You should provide a response that helps the user answer the question.
Keep the answer as concise as possible; single-sentence answers are best.
Your answer must be in JSON formatted dictionary with the key "answer".
Generate json-formatted dictionary only, do not include any other information like formatting, etc.
example:
{"answer": "Gothic architecture"}
\end{lstlisting}
\end{lstfloat}

\begin{lstfloat}[h]
\begin{lstlisting}[label={lst:prompt_eval},language={},caption={Full example of our prompt given to GPT-4 for answer evaluation.}]
You are a helpful assistant to a person who is trying to answer the question about an image from a examiner.
The person has provided a reasoning process to reach the final answer, and then provide a final answer.
Your task is to evaluate the answer in score 1-4.
1: The answer is wrong.
2: The answer is somewhat correct; meaning of the answer is similar to the correct answer but details are wrong.
3: The answer is mostly correct; meaning of the answer is correct but some details are wrong.
4: The answer is completely correct.
Before you evaluate the answer, you need to extract the final answer from the user's full response.
When extracting the answer, you should not modify the user's wording; just extract the answer as it is.

Your answer must be in JSON formatted dictionary with the key "answer" (str) and "score" (int).
Generate json-formatted dictionary only, do not include any other information like formatting, etc.
example:
{"answer": "Gothic architecture", "score": 4}
{"answer": "Cathedral", "score": 2}
{"answer": "School", "score": 1}
{"answer": "Baroque architecture", "score": 3}
\end{lstlisting}
\end{lstfloat}

Here, we provide the full example of the prompt given to GPT-4 for answer generation and evaluation.
The prompt for answer generation is shown in Listing~\ref{lst:prompt_answer}, and the prompt for answer evaluation is shown in Listing~\ref{lst:prompt_eval}.
In both prompts, we provide an instruction and several examples to guide GPT-4 to generate and evaluate the answer, respectively.
We instructed GPT-4 to return the answer in JSON format, which enables us to parse the answer and evaluate it automatically.

\subsection*{B. More Examples of the Dataset}\label{sec:appendix_data}

\begin{figure*}[b]
    \centering
    \includegraphics[width=0.94\linewidth]{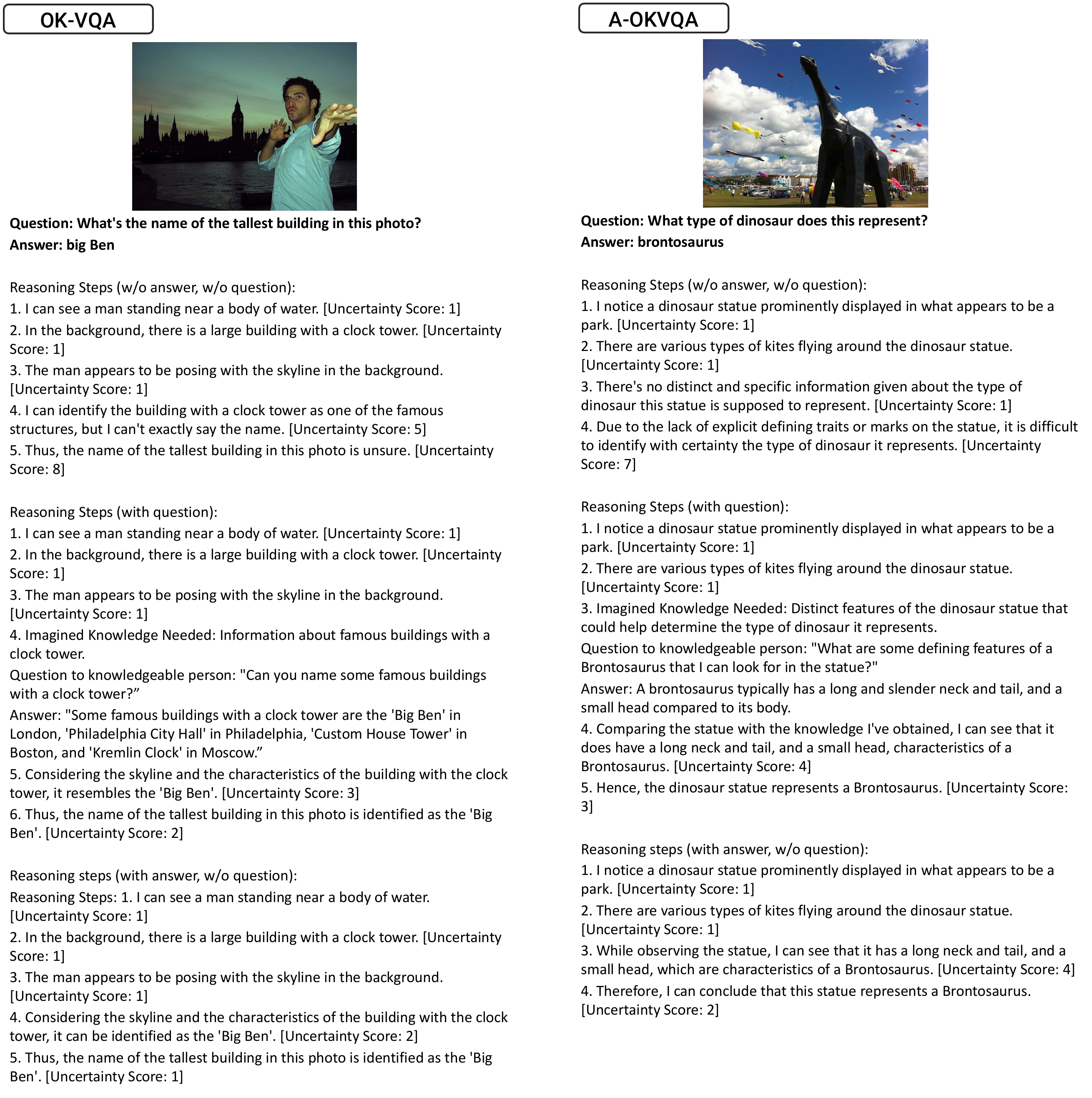}
    \caption{Examples from our dataset, based on OK-VQA and A-OKVQA.}
    \label{fig:app_data1}
\end{figure*}

\begin{figure*}
    \centering
    \includegraphics[width=\linewidth]{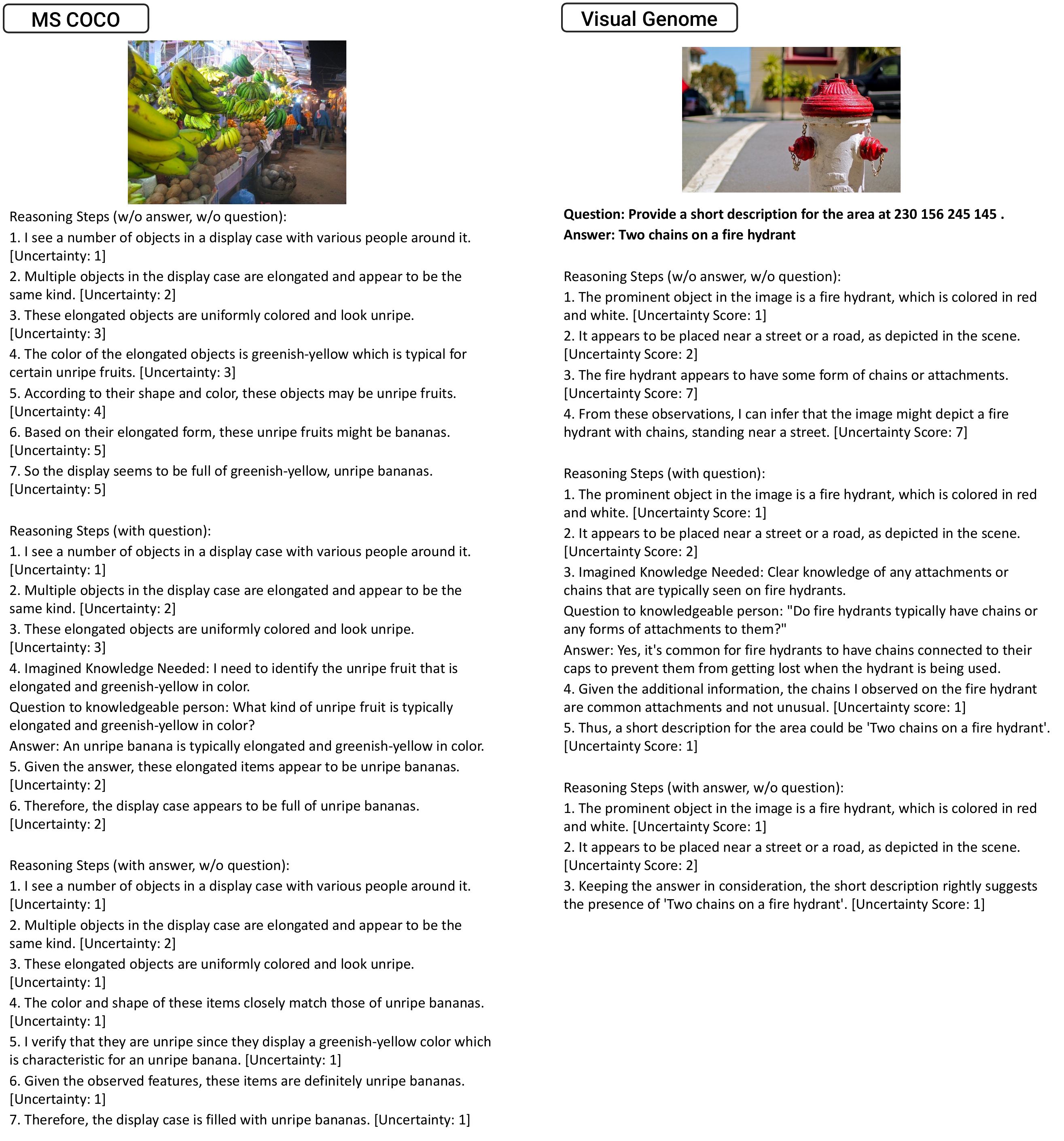}
    \caption{Examples from our dataset, based on MS COCO caption and Visual Genome caption.}
    \label{fig:app_data2}
\end{figure*}

In Figure~\ref{fig:app_data1} and~\ref{fig:app_data2}, we show more examples of the dataset we crafted based on different datasets: OK-VQA, A-OKVQA, MS COCO caption, Visual Genome caption, OVEN, and Encyclopedic-VQA.
Each example consists of an image, a question, and a reasoning chain that explains how to answer the question or generate a caption for the image.
We provide the reasoning chain of three types: (1) reasoning steps without answer and question, in which the data is generated without access to the ground-truth answer and there is no question generation step; (2) reasoning steps with question, in which the data is generated with access to the ground-truth answer and there is a question generation step; and (3) reasoning steps with answer and without question, same as the second type but without the question generation step.

\subsection*{C. More Qualitative Results}\label{sec:appendix_qualitative}

\begin{figure*}
\centering
\includegraphics[height=0.9\textheight]{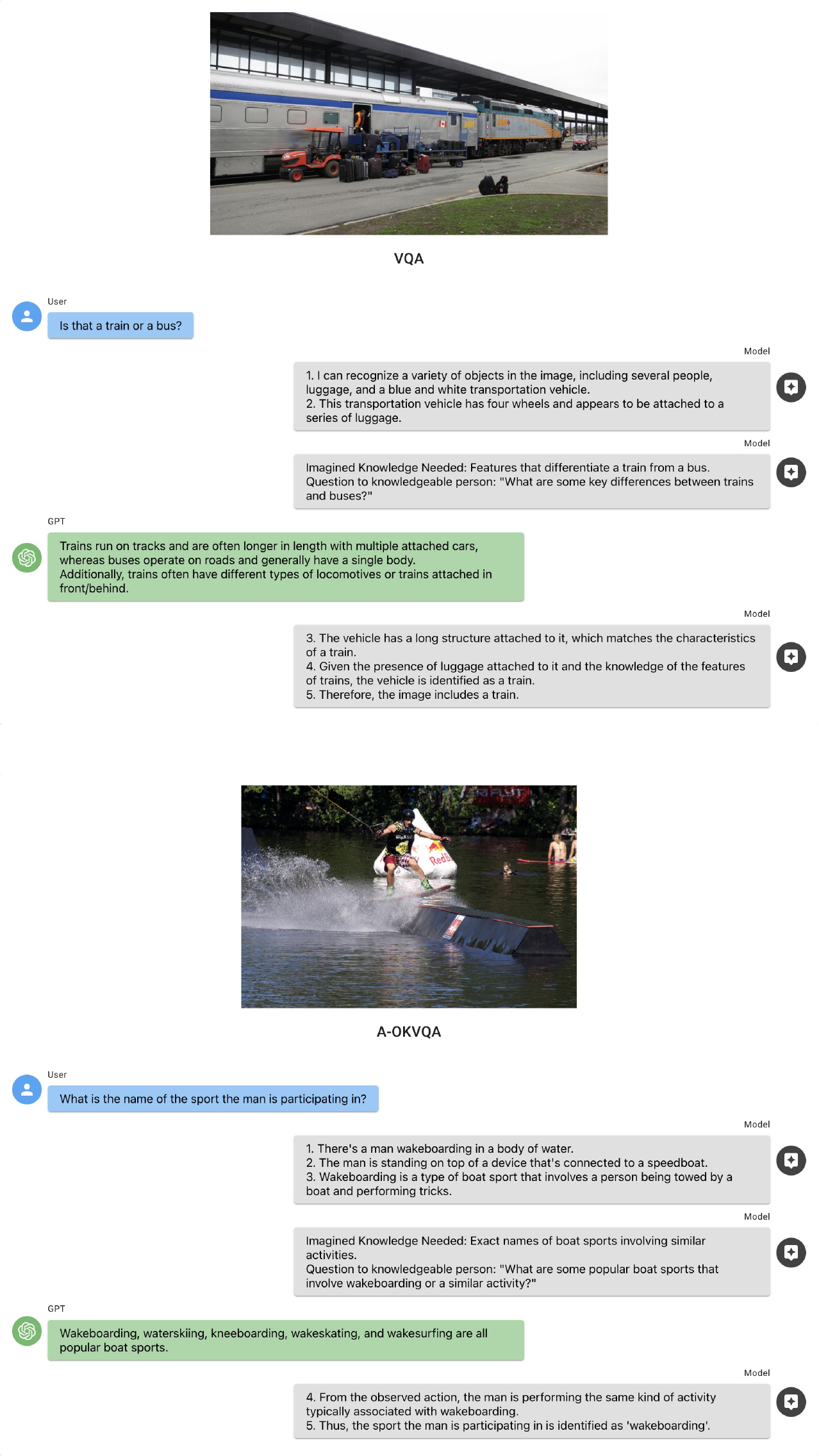}
\caption{Additional qualitative results on generating reasoning steps for VQA and A-OKVQA caption.}
\label{fig:app_result1}
\end{figure*}

\begin{figure*}
    \centering
    \includegraphics[height=0.9\textheight]{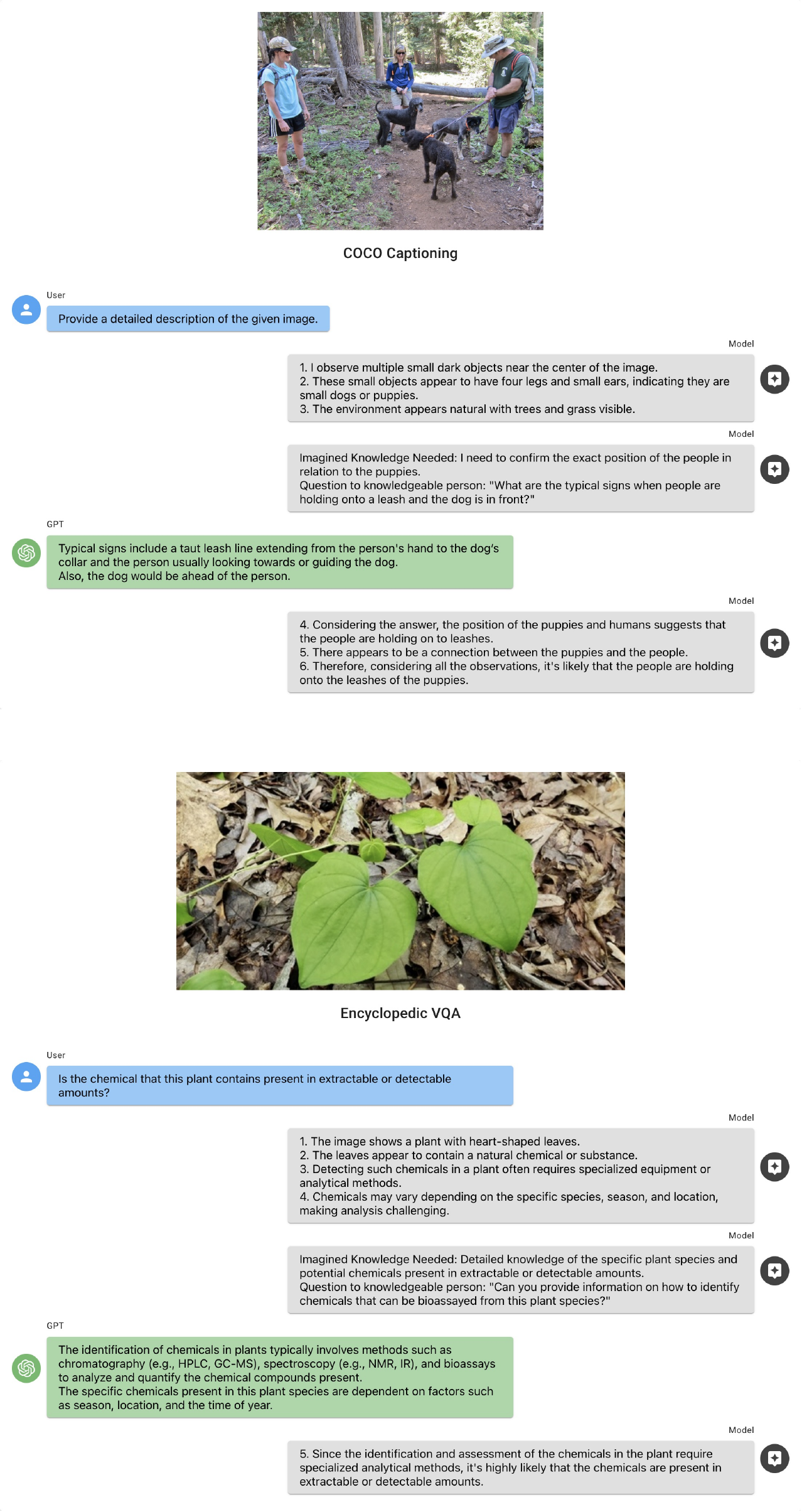}
    \caption{Additional qualitative results on generating reasoning steps for MS COCO caption and Encyclopedic VQA.}
    \label{fig:app_result2}
\end{figure*}

\begin{figure*}
    \centering
    \includegraphics[height=0.9\textheight]{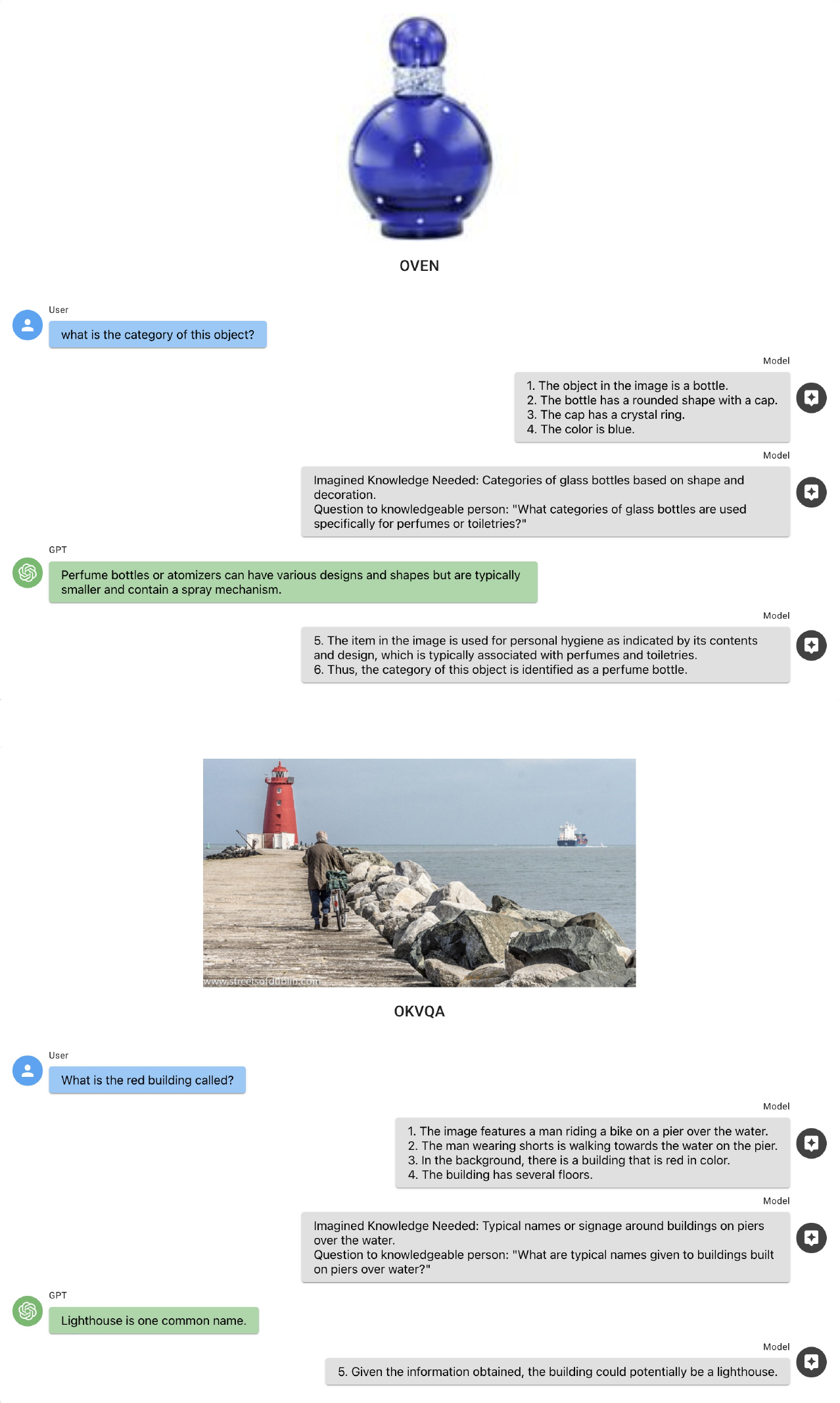}
    \caption{Additional qualitative results on generating reasoning steps for OVEN and OK-VQA.}
    \label{fig:app_result3}
\end{figure*}

\begin{figure*}
    \centering
    \includegraphics[height=0.45\textheight]{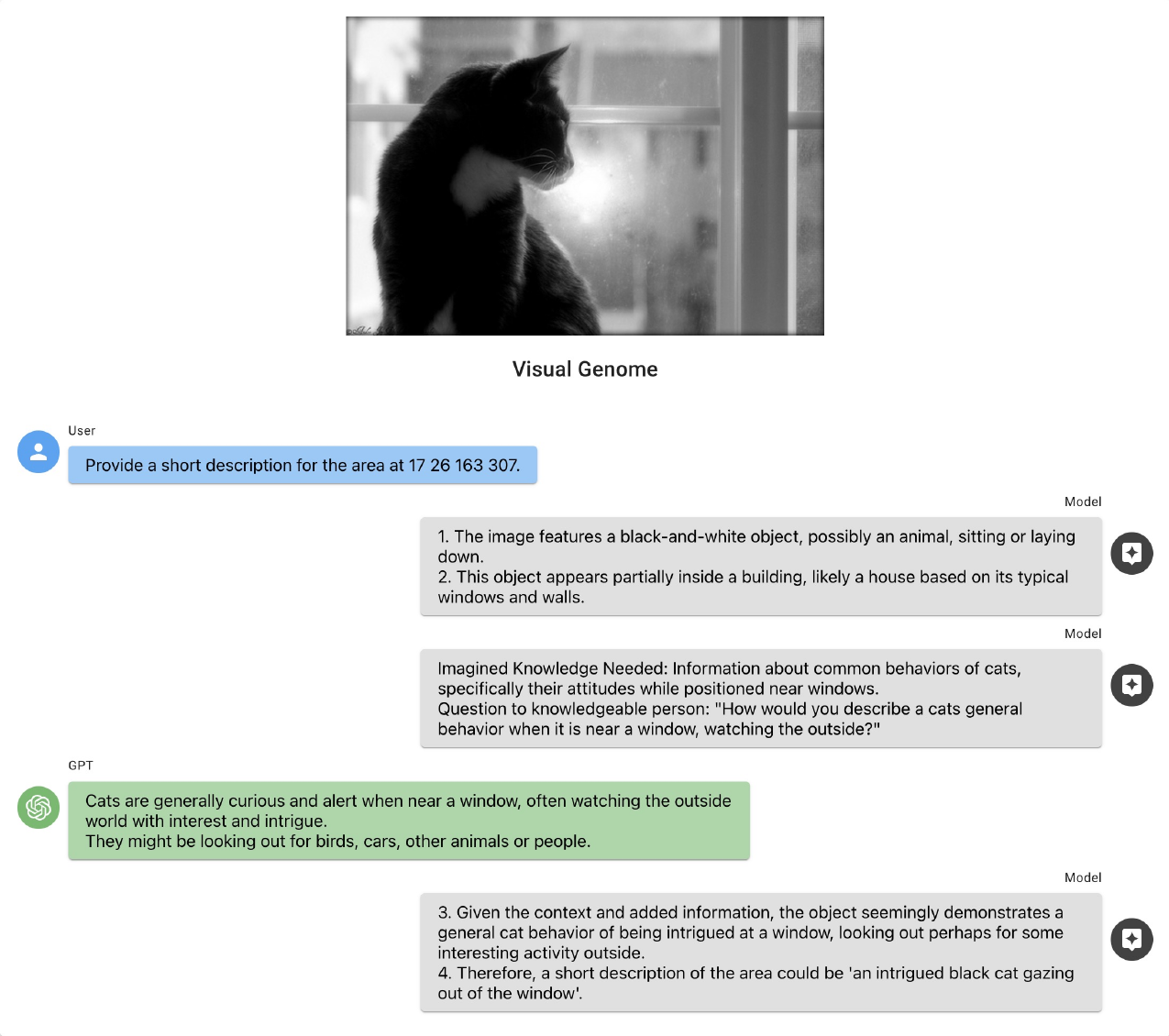}
    \caption{Additional qualitative results on generating reasoning steps for Visual Genome.}
    \label{fig:app_result4}
\end{figure*}

We show more qualitative results of generated reasoning steps by our model for VQA, A-OKVQA, COCO Caption, Encyclopedic VQA, OVEN, OK-VQA, and Visual Genome in Figure~\ref{fig:app_result1}, ~\ref{fig:app_result2}, ~\ref{fig:app_result3}, and ~\ref{fig:app_result4}.

\begin{figure*}
    \centering
    \includegraphics[width=\linewidth]{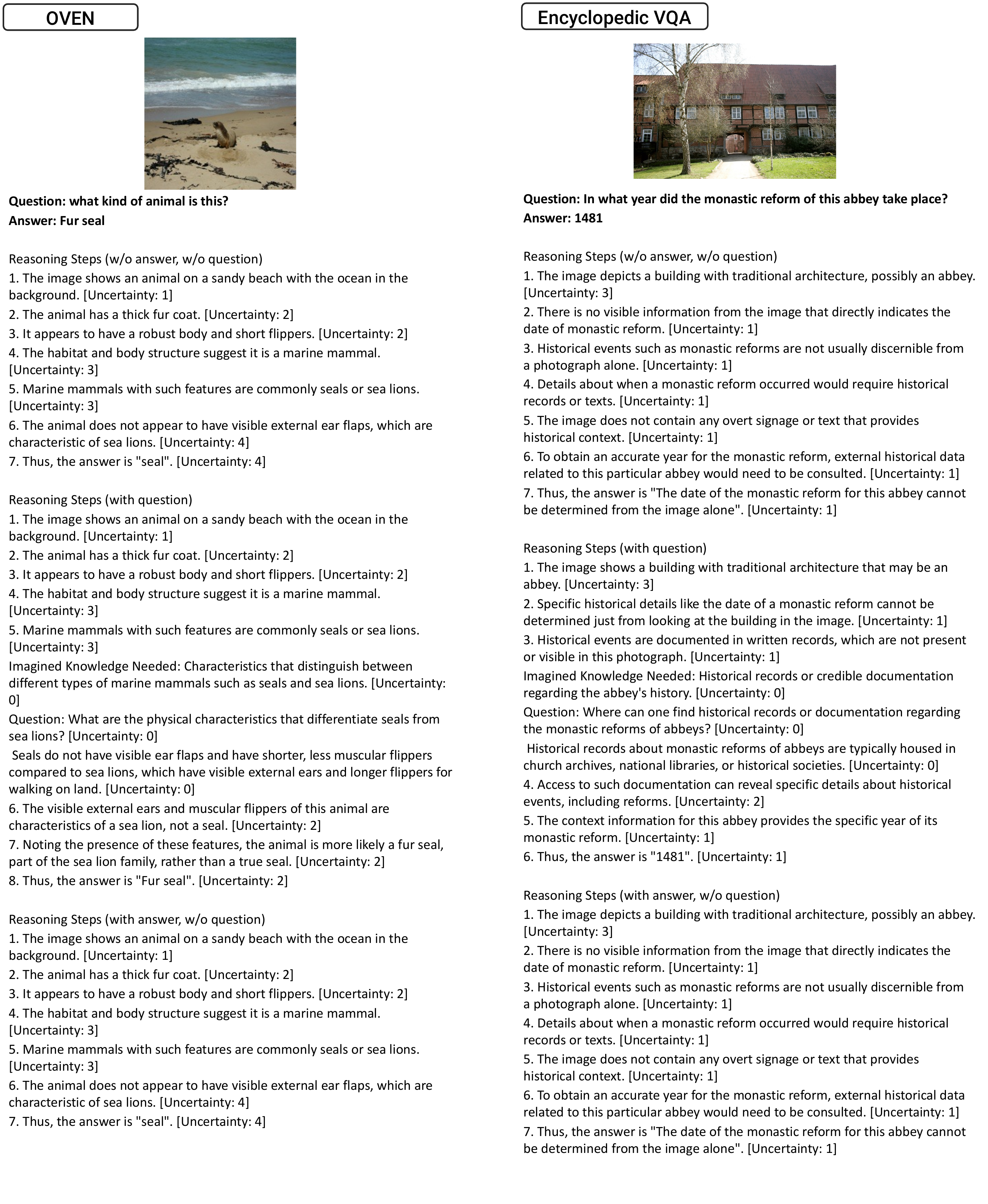}
    \caption{Examples from our dataset, based on OVEN and Encyclopedic-VQA.}
    \label{fig:app_data3}
\end{figure*}

\end{document}